\documentclass[journal]{IEEEtran}
\usepackage[numbers,sort&compress]{natbib}
\usepackage{amsfonts}
\usepackage[cmex10]{amsmath}
\usepackage{array}
\usepackage{mdwmath}
\usepackage{mdwtab}
\usepackage{eqparbox}
\usepackage[font=normalsize]{caption}
\usepackage{hyperref}
\usepackage{stfloats}
\usepackage{url}
\usepackage{amssymb}
\usepackage{float}
\usepackage{indentfirst}
\usepackage{makeidx}
\usepackage{tabularx}
\usepackage{cases}
\usepackage{multirow}
\usepackage{booktabs}
\usepackage{diagbox}
\usepackage{mathtools}
\usepackage{color}
\usepackage{stfloats}
\usepackage{lipsum}
\usepackage{amsfonts}
\usepackage{bm}
\usepackage{dsfont}
\usepackage{lipsum}
\usepackage{amsthm}
\usepackage{graphicx, epsfig, subfigure}
\usepackage[font=small]{caption}
\allowdisplaybreaks[4]

\makeatletter

\newif\if@restonecol
\makeatother

\usepackage[linesnumbered,ruled,vlined]{algorithm2e}%[ruled,vlined]{
\usepackage{algpseudocode}
\usepackage{amsmath, bm}
\begin{document}
\title{Communication-Efficient Personalized Distributed Learning with Data and Node Heterogeneity}
\author{Zhuojun~Tian,~\textit{Member, IEEE,}
        Zhaoyang~Zhang,~\textit{Senior Member, IEEE,}
        Yiwei~Li,~\textit{Senior Member, IEEE,}\\
        and Mehdi~Bennis,~\textit{Fellow, IEEE}
        \thanks{
        This work was supported in part by National Natural Science Foundation of China under Grants 62394292 and U20A20158, Ministry of Industry and Information Technology under Grant TC220H07E, Zhejiang Provincial Key R\&D Program under Grant 2023C01021, the Fundamental Research Funds for the Central Universities No. 226-2024-00069, and the EU-SNS 6G CENTRIC Project.}
        \thanks{
        Z.~Tian (email: dankotian@zju.edu.cn) was with the College of Information Science and Electronic Engineering, Zhejiang University, Hangzhou 310027, China and now is with the Center for Wireless Communications, University of Oulu, Oulu 90014, Finland. 
        
        Z.~Zhang (Corresponding Author, email: ning\_ming@zju.edu.cn) is with the College of Information Science and Electronic Engineering, Zhejiang University, Hangzhou 310027, China, and with the State Key Laboratory of Industrial Control Technology, Hangzhou 310027, China, and also with Zhejiang Provincial Key Laboratory of Multimodal Communication Networks and Intelligent Information Processing, Hangzhou 310027, China.
        
        Y. ~Li (e-mail: lywei0306@foxmail.com) is with Fujian Key Laboratory of Communication Network and Information Processing, Xiamen University of Technology, Xiamen 361024, China.
        
        M.~Bennis (e-mail: mehdi.bennis@oulu.fi) is with the Center for Wireless Communications, University of Oulu, Oulu 90014, Finland.
        }
        \thanks{
        The code for the paper is available on \href{ https://github.com/ZhuoJTian/MCE-PL}{https://github.com/ZhuoJTian/MCE-PL}.}
}

\maketitle
	
\begin{abstract}
To jointly tackle the challenges of data and node heterogeneity in decentralized learning, we propose a distributed strong lottery ticket hypothesis (DSLTH), based on which a communication-efficient personalized learning algorithm is developed.
In the proposed method, each local model is represented as the ‌Hadamard product of global real-valued parameters and a personalized binary mask for pruning.  
The local model is learned by updating and fusing the personalized binary masks while the real-valued parameters are fixed among different agents. 
To further reduce the complexity of hardware implementation, we incorporate a group sparse regularization term in the loss function, enabling the learned local model to achieve structured sparsity.
Then, a binary mask aggregation algorithm is designed by introducing an intermediate aggregation tensor and adding a personalized fine-tuning step in each iteration, which constrains model updates towards the local data distribution. 
The proposed method effectively leverages the relativity among agents while meeting personalized requirements in heterogeneous node conditions. 
We also provide a theoretical proof for the DSLTH, establishing it as the foundation of the proposed method. 
Numerical simulations confirm the validity of the DSLTH and demonstrate the effectiveness of the proposed algorithm. %The results show that the algorithm can collaboratively learn personalized models with high accuracy while incurring minimal communication costs. 
\end{abstract}

\section{Introduction} \label{sec1}
	As one of the most promising applications in 6G era, Artificial Intelligence of Things (AIoT) combines the artificial intelligence technologies with the Internet of Things (IoT) infrastructure, resembling the transformation from ``connected things" to ``connected intelligence" .  In AIoT, edge learning has been envisioned as the key enabler to embed model training and inference into network, utilizing the computation and communication capability of devices. In conventional edge learning procedures, each agent has access to its own training data and cooperates with others to obtain a common global model~\cite{liprivate2024}. However, due to different geographical locations or various users to serve,  agents often have partial views of global information \cite{Tian2023D}, leading to heterogeneous data distributions. Under such non-identically and independently distributed (non-i.i.d.) conditions, the consensus model typically performs poorly on locally accessible data, motivating researchers to explore personalized learning among agents~\cite{tian2023}. 
	
	Recently, in the increasingly complex AIoT application scenarios, agents tend to exhibit diverse system capabilities, ranging from servers with high computational power and large storage capacity to mobile phones or sensors with limited capabilities.  In this sense, the concept of personalization needs to be expanded to address  the challenge of node heterogeneity, where each agent's personalized model should adapt to local system characteristics. This {motivates} us to jointly consider the data (statistical) and node (system) heterogeneity in AIoT system by developing personalized models for each agent. Moreover, the frequent communication in the training process brings  high communication costs, which is a major challenge in AIoT system. {Therefore, our goal is to design a communication-efficient personalized algorithm that accounts for both node and data heterogeneity.}

	In this work, we investigate decentralized communication networks. Different from the popular Federated Learning (FL) framework, decentralized learning does not rely on a central node to collect and process all agents' messages, thereby alleviating the heavy communication burden on the center node and enhancing robustness against nodes' failure. In decentralized learning,   each agent typically updates its local model and then shares the updated model with neighboring nodes for aggregation. The standard aggregation methods, such as averaging or weighted averaging used in decentralized stochastic gradient descent~\cite{Lian2017}, often perform poorly in non-i.i.d. conditions. Therefore, a specialized aggregation design is necessary to effectively leverage the relationships among agents in this non-i.i.d. and heterogeneous system.  

     \subsection{Related Works}
	
    {Recently, many works have focused on addressing the non-i.i.d. challenges in FL by designing personalized learning algorithms \cite{ Caldarola2021, li2020federated, Fallah2020, Ghosh2020, Dinh2020}. Specifically, meta-learning methods have been applied  in FL \cite{Fallah2020} to enable personalized training. } The work~\cite{Caldarola2021} clustered the agents and used the graph convolution networks to share knowledge across different clusters. More recently, the node heterogeneity problem in FL has gained attention, particularly where clients have distinct capabilities \cite{diao2020heterofl}. The existing methods can generally  be classified into three categories: split-learning based methods \cite{He2020, Gupta2018}, submodel training methods \cite{diao2020heterofl, horvath2021fjord, deng2022tailorfl, Seo2021communication, jiang2022model} and the methods based on factorization \cite{yao2021fedhm, niu2022federated, mei2022resource}. The work~\cite{diao2020heterofl} questioned the assumption of the same global model and proposed the HeteroFL framework, where local model parameters in each client form a subset of the global model, and aggregation is performed through partial averaging. The work~\cite{horvath2021fjord} introduced ordered pruning to extract and train different submodels for different clients. Other works~\cite{yao2021fedhm, niu2022federated, mei2022resource} focused on compressing large global models into small models by low-rank factorization and assign the small models to different clients based on their capabilities. In all these methods, a central server is responsible for maintaining the large global model and handling tasks such as model partition, compression and factorization. However, in the decentralized framework, there is no central processor to perform these tasks, making it challenging to directly apply these methods to decentralized learning.  
	
	On the other hand, there have been fewer works addressing personalization in decentralized learning \cite{tian2023, zantedeschi2020fully, chen2024enhancing, xiong2024deprl}. The work~\cite{zantedeschi2020fully} constructed a collaborative graph to represent the correlation among the tasks of different agents. During  training, both the collaborative graph and the local models are alternatively learned to obtain the personalized model for each agent, which however highly increases the computation and communication cost. The methods in \cite{tian2023} and \cite{chen2024enhancing} utilize  partially shared local models. The work~\cite{tian2023} employed a graph attention mechanism to jointly learn aggregation weights, while the work~\cite{chen2024enhancing} introduced a topology reconstruction algorithm to reduce the communication bandwidth and accelerate the training process. Nevertheless,  the above works only focused on  statistical personalization, where the personalized models share a common neural network architecture, without addressing node heterogeneity. 
    {To resolve this issue, the work~\cite{dai2022dispfl} proposed the Dis-PFL protocol employing personalized sparse masks, where the active parameters and local masks are transmitted while those intersection weights are averaged.   The work~\cite{hong2021dlion} proposed dynamically adjusting batch sizes to enhance data parallelization and exchanging prioritized gradients to reduce communication overhead. Meanwhile, the work~\cite{yang2020mitigating} addressed the long-tail effects caused by node heterogeneity, introducing a hierarchical synchronization procedure to minimize latency. }

	In distributed learning, the large number of transmitted model parameters and the continuous iterations lead to high communication costs in practical wireless communication systems. Consequently, there have been extensive works aiming to reduce the communication cost through resource allocation, over-the-air computation \cite{yang2020federated}, accelerating convergence \cite{wang2019adaptive}, and minimizing  communication cost per iteration \cite{li2021communication, lincns2024, doostmohammadian2024nonlinear, doostmohammadian2025log}. In the latter approach, some works have explored client selection while others have focused on  model compression by model sparsification, gradient compression or parameter quantization. {The authors in \cite{doostmohammadian2024nonlinear} proposed a first-order computationally efficient distributed optimization algorithm, which shows stability in time-varying networks, and they extend the methods in \cite{doostmohammadian2025log} for log-scale quantized data exchange.}
    These techniques can be generally applied in both FL and decentralized frameworks. 
    {Additionally, the authors in \cite{fotohi2024lightweight} introduced the communication-efficient FL techniques using compression, which meanwhile solves privacy concerns against poisoning and inference attacks through secure aggregation. The secure blockchain-enabled algorithm in \cite{fotohi2024lightweight} further utilizes blockchain-enable steps to address the privacy issue.}
    
    The work~\cite{Li2021Fedmask} proposed to reduce the communication costs by utilizing the lottery ticket hypothesis, where the sparse binary masks are learned and transmitted among agents. The masks from different clients are partially aggregated, where overlapping  entries across the binary masks are aggregated while non-overlapping entries remain unchanged. This approach effectively learns personalized models for each agent while significantly reducing communication costs. Similar to the procedure in \cite{Li2021Fedmask}, many works \cite{li2020lotteryfl, Seo2021communication, jiang2022model, dai2022dispfl} have applied the lottery ticket hypothesis as a pruning technique to tackle the heterogeneity in distributed learning. The authors in \cite{li2020lotteryfl} proposed LotteryFL for data heterogeneity, where clients prune their local model based on the received global model. Concerning the scalability and stragglers problem in LotteryFL, another algorithm is designed in \cite{Seo2021communication} exploiting the downlink broadcast. The authors in \cite{jiang2022model} further considered the system heterogeneity and designed a two-stage pruning method on both client and server sides, where the model size is adapted to reduce both communication and computation overhead. However, it sill requires the central server to prune all clients. To provide a good solution for decentralized heterogeneity, the authors in \cite{dai2022dispfl} proposed Dis-PFL to save the communication and computation cost. However, in all of the above algorithms, the real-valued model parameters are transmitted, leading to high communication cost even with sparse masks.

 \subsection{Contributions}
    {To further reduce communication costs in decentralized learning framework that simultaneously considers data and node heterogeneity, we adopt the core idea from \cite{Li2021Fedmask} to learn personalized sparse binary tensors.} We propose and verify the distributed strong lottery ticket hypothesis (DSLTH), serving as the theoretical foundation for this approach.
    In \cite{Li2021Fedmask}, the initialization of the real-valued mask from the aggregated binary tensor is challenging. Moreover, the model aggregation based on overlapping masks is unsuitable for system heterogeneous scenarios, as the aggregated mask tends to become sparser, limiting the utilization of the diverse capabilities of agents. To solve this issue, we design a novel aggregation algorithm for the binary tensors. Note that our work differs from \cite{dai2022dispfl}, where the proposed method only transmits binary masks among agents and the personalized masks are collaboratively learned. On the contrary, Dis-PFL in \cite{dai2022dispfl} transmits both real-valued parameters and masks and updates both of them. {Compared with the methods based on quantization \cite{doostmohammadian2025log}, the mask is naturally binary, reducing the communication cost without quantization.}
    
    Specifically, we integrate the information from neighboring binary masks into the local real-valued mask tensor through an intermediate aggregation tensor, where the influence of the binary information is adaptively modulated by our specialized design. Moreover, we incorporate a personalized fine-tuning step in each iteration to ensure that the learned personalized model aligns with the local data distribution. Unlike methods that fuse binary masks through intersection, our algorithm aggregates binary mask information directly into real-valued tensors, offering greater flexibility for heterogeneous conditions. Our key contributions can be summarized as follows:
    \begin{itemize}
        \item{We propose DSLTH for heterogeneous distributed learning systems, laying the foundation of the algorithm development. By introducing an intermediate aggregation tensor, the binary masks from neighboring nodes can be fused into local real-valued mask tensor. Moreover, we incorporate a personalized fine-tuning step in the update process, ensuring that the personalized model moves towards local data distribution.}
        
        \item{This work presents a novel framework for decentralized learning that simultaneously addresses both data and node heterogeneity by tailoring personalized models for individual agents. Additionally, based on the proposed DSLTH, the communication cost is significantly reduced during the training process while guaranteeing the learning performance, making this approach highly applicable for future AIoT systems.}
        
        \item{The distributed strong lottery ticket hypothesis is theoretically proved for non-i.i.d. data distribution. We further verify the hypothesis in node heterogeneity through testing experiments. The effectiveness and communication-efficiency of the proposed algorithm are demonstrated through experimental results, where the agents collaboratively learn their personalized and heterogeneous models with minimal  communication cost.}
\end{itemize}
	
	The rest of this article is organized as follows: Section \ref{pre} describes the system model and the model learning method based on binary masks. In Section \ref{algo}-A,  we propose the distributed strong lottery ticket hypothesis. 
	Then in Section \ref{algo}-B, we introduce the group sparsity regularization term added in the loss function and then describe the rule to obtain the binary mask tensor.
    Then in Section \ref{algo}-C, we develop the aggregation and updating procedure named as MCE-PL for communication-efficient personalized learning. 
    Section \ref{Sec4} provides the theoretical proof for the proposed DSLTH, providing the theoretical foundation of the proposed method.
    Simulation results are represented in Section \ref{experiment}, followed by the conclusion in Section \ref{conclusion}.
	
\section{Preliminary}\label{pre}
\subsection{System Model}
\begin{figure}[!htp]
    \centering
    \includegraphics[width=0.45\textwidth]{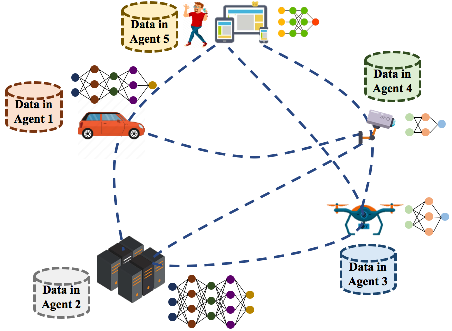}
    \caption{The decentralized communication network of agents with heterogeneous data and system capabilities.}
    \label{fig1}
\end{figure}

Consider a multi-agent decentralized communication network as shown in Fig. \ref{fig1}, which can be represented by an undirected graph $\mathcal{G}=(\mathcal{V},\mathcal{E})$. {The communication network is assumed to be connected and static over the training process.}
In $\mathcal{G}$, $\mathcal{V}=\{1,...,N\}$ denotes the set of $N$ distributed agents and $\mathcal{E}=\{\varepsilon_{ij}\}_{i,j \in \mathcal{V}} $ represents the set of communication links between any two adjacent agents. Let $\mathcal{N}_i$ denote the set of all neighboring agents connected with agent $i$ and we denote the number of agents in $\mathcal{N}_i$ by $d_i=|\mathcal{N}_i|$. The adjacency matrix of $\mathcal{G}$ is denoted by $\textbf{A}$, where $\textbf{A}(i,j)=1$ if $\varepsilon_{ij}\in\mathcal{E}$ and $\textbf{A}(i,j)=0$ otherwise. 

As shown in Fig. \ref{fig1}, the agents in the network are located in different positions within a global environment, leading to each agent having only a partial view of the global information.
Thus, each agent $i\in\mathcal{V}$ has access to its own local training dataset $\mathcal{D}_i=\{\bm{x}_s, \bm{y}_s\}_{s=1}^{n_i}$, with data drawn from a personal data distribution over a  common feature space $\mathcal{X}$ and label space $\mathcal{Y}$. 
In the considered model, the agents in $\mathcal{G}$ have different data distributions, known as statistical heterogeneity. 
Additionally,  agents may  possess varying system capacities, including differences in computational power or communication capabilities, resulting in system heterogeneity among agents. 
We define the model retention ratio as the proportion of model that needs to be retained after pruning,  denoted by $0<r_i\le1$ for agent $i$. 
Then the model sparsity requirement in agent $i$ is represented by $1-r_i$. 
The model retention ratio indicates the system capacity of  each agent, where a higher $r_i$ value corresponds to greater communication or computational capacity. 

Let $f_i$ denote the local loss function of agent $i$ and the global loss function is:
\begin{equation}
	\label{opt_sys}
	\min F(\textbf{V}) := \frac{1}{N}\sum\nolimits_{i=1}^{N}f_i(\bm{v}_i),
\end{equation}
where $\bm{v}_i$ denotes the model parameters of node $i$, and $\textbf{V} = [\bm{v}_1, \bm{v}_2, \ldots, \bm{v}_N] $ represents the collection of model parameters across all nodes.
In supervised learning, $f_i(\bm{v}_i)$ stands for the expected loss over the local data distribution of agent $i$, defined by 
\begin{align}
f_i(\bm{v}_i) := \mathbb{E}_{\mathcal{D}_i}[l_i(\bm{v}_i; \bm{x}_s, \bm{y}_s)],
\end{align}
where $l_i(\bm{v}_i; \bm{x}_s, \bm{y}_s)$ measures the prediction  error for label $\bm{y}_s$ given the input $\bm{x}_s$ and the model parameters $\bm{v}_i, \forall i \in \mathcal{V}$. 
We decouple the local model parameters $\bm{v}_i$ into the Hadamard product of real-valued parameters $\bm{w}_i$ and binary masks $\bm{m}_i$, i.e., $\bm{v}_i=\bm{w}_i\odot \bm{m}_i, \forall i\in\mathcal{V}$. The binary mask $\bm{m}_i$ has the same shape as the model parameters, with each entry being $0$ or $1$.
Due to the statistical and system heterogeneity among agents, this personalized learning task's, optimization problem \eqref{opt_sys} does not include consensus constraints on the model parameters. {Moreover, $f_i$ is assumed to be $L$-smooth.}
{In this work, we consider the synchronous settings, where each agent aggregates the information until it receives all the messages from neighboring nodes. This may require more training time, due to the heterogeneous communication or learning delays. However, such synchronous settings can improve the stability of the algorithm.}

Before developing our algorithm, we introduce the lottery ticket hypothesis and the model learning procedure based on binary masks.

\subsection{Lottery Ticket and Binary Mask Update}\label{sec2_b}
\begin{figure*}[!htp]
    \centering
    \includegraphics[width=0.95\textwidth]{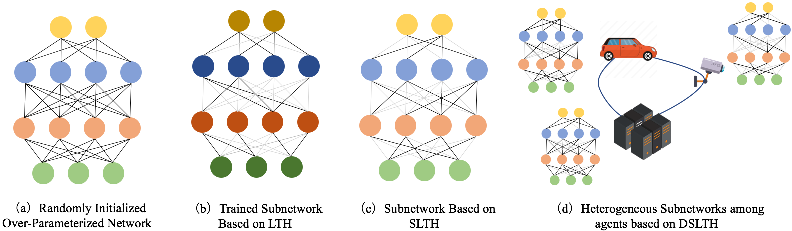}
    \caption{Illustration of the lottery ticket hypothesis, strong lottery ticket hypothesis and distributed strong lottery ticket hypothesis:  {In (a), (c) and (d), the weight parameters are of initialized values, while the circles of darker color in (b) indicate trained weight parameters.}}
    \label{sys2}
    \vspace{-0.5cm}
\end{figure*}

The lottery ticket hypothesis (LTH) was proposed and experimentally verified in \cite{frankle2018lottery} stating that any randomly initialized network contains lottery tickets. Here the lottery tickets are sparse subnetworks that can be trained to achieve the performance of the fully-trained original network, as shown in Fig. \ref{sys2}(b).
More recently, the work~\cite{ramanujan2020} proposed a stronger version of the hypothesis,    depicted in Fig. \ref{sys2}(c): a network with random weights contains sub-networks that can approximate any given sufficiently-smaller neural networks with high probability, even without training. This concept, termed the Strong Lottery Ticket Hypothesis (SLTH) in \cite{pensia2020optimal}, has been rigorously proven for dense networks~\cite{malach2020proving} and convolutional neural networks~\cite{da2022proving}.

However, it is always computationally challenging to find any winning lottery tickets with competitive performance on the given data. 
The work~\cite{zhou2019deconstructing} demonstrated  the combination of pruning mask and weights can create an efficient subnetwork found within the larger network, while the work~\cite{Li2021Fedmask} proposed an effective way that involves updating binary masking tensor while keeping real-valued parameters fixed.

Specifically, since the binary tensor $\bm{m}_i$ is discrete and cannot be directly updated using gradient descent, we introduce  a real-valued mask tensor $\bm{z}_i$ with the same size as $\bm{m}_i$. 
By updating $\bm{z}_i$ and ranking the absolute values of its entries, each binary entry in $\bm{m}_i$ can be determined as $\bm{m}_i(t)=\text{Thres}(\text{abs}(\bm{z}_i(t)))$, where $\text{Thres}(\cdot)$ is an entry-wise threshold operation defined as:
\begin{equation}
    \label{thres}
    \text{Thres}({a})=\left\{
        \begin{array}{rcl}
        0       &      & {{a}\le T_a,}\\
        1     &      & {{a}>T_a.}
        \end{array} \right. 
\end{equation}
Here $T_a$ is chosen according to the required model retention ratio. 
Considering the different scale among layers in deep neural network, the threshold operation is conducted in a layer-wise way. 
For the real-valued mask tensor in layer $l$, denoted as $\bm{z}_{i,l}$, we rank all entries from largest to smallest, and select the $100\times r_i$-th entry as $T_a$. In this way, the entries of $1$ in the binary mask tensor $\bm{m}_{i,l}$ correspond to the $r_i$ largest amplitudes in $\bm{z}_{i,l}$. 
Due to the system heterogeneity, the value of $r_i$ may differ among agents, leading to different sparsity of the binary mask tensors.

In the $k$-th iteration, according to the back propagation, we can get the approximated update rule of $\bm{z}_i$ based on gradient descent:
\begin{equation}\label{mask_update_z}
    \begin{aligned}
        \bm{z}_i^{(k)} & = \bm{z}_i^{(k-1)}-\eta\frac{\nabla f_i}{\nabla \bm{v}_i}\times \frac{\nabla \bm{v}_i}{\nabla \bm{m}_i}\times \frac{\nabla \bm{m}_i}{\nabla \bm{z}_i}\\
        & \approx \bm{z}_i^{(k-1)}-\eta\frac{\nabla f_i}{\nabla \bm{v}_i}\times \frac{\nabla \bm{v}_i}{\nabla \bm{m}_i}\times \text{sign}(\bm{z}_i^{(k-1)}),
    \end{aligned}
\end{equation}
where the gradient ${\nabla \bm{m}_i}/{\nabla \bm{z}_i}$ cannot be explicitly calculated, so we approximate it using $\text{sign}(\bm{z}_i^{(k-1)})$, a method also applied in the training of binary neural networks. 
In each iteration, parameters of the real-valued mask tensor $\bm{z}_i$ are updated according to \eqref{mask_update_z}. Subsequently, the binary mask tensor $\bm{m}_i$  is obtained through the threshold operation defined in \eqref{thres}.

\section{Algorithm Development}\label{algo}

In this section, we firstly propose the distributed strong lottery ticket hypothesis (DSLTH), which serves as the theoretical foundation for the proposed method. {The proposed DSLTH justifies the designed framework where only binary masks are transmitted and updated among agents.
Next, we introduce a regularization term with group sparsity and design the procedure for obtaining the binary mask tensor. 
Finally,  the limited information embedded in these binary mask tensors makes it challenging to effectively fuse the mask tensors from neighboring, particularly in scenarios with node heterogeneity.} To resolve this issue, we develop a distributed training algorithm tailored for the considered scenario and give the Mask-based Communication-Efficient Personalized Learning Algorithm (MCE-PL).

\subsection{Distributed Lottery Ticket Hypothesis}\label{sec3_a}
According to the strong lottery ticket hypothesis, there exist subnetworks within a large neural network that can achieve roughly the comparable accuracy as the target network.
At the same time, recent studies in FL~\cite{diao2020heterofl, horvath2021fjord, Li2021Fedmask} have demonstrated that,  under the non-i.i.d. condition and the diverse system capabilities of agents, different extraction of subnetworks from a large, well-trained model can achieve high personalized accuracy. 
Building on these observations, we propose a DSLTH  for distributed learning systems as follows and shown in Fig. \ref{sys2}(d). 

\textit{\textbf{Distributed strong lottery ticket hypothesis:} In distributed learning scenarios  characterized by both data and node heterogeneity, there exist heterogeneous subnetworks among agents  that can be extracted from a global and randomly initialized large (over-parameterized) neural network, performing well on local data distribution with high probability.}

Note that the rigorous proof of DSLTH in general conditions is as challenging as proving LTH and SLTH~\cite{pensia2020optimal, malach2020proving, da2022proving}.
Therefore, in this work, we present a theoretical proof of DSLTH under constrained CNN conditions in Section \ref{Sec4}. Additionally, we empirically verify the hypothesis in more general settings through testing simulations, as described in Section \ref{veri_DSLTH}. We leave a more comprehensive evaluation, both experimental and theoretical, to future works.

Based on the proposed DSLTH, the real-valued parameters $\bm{w}_i$ are initialized and fixed the same among agents $i\in\mathcal{V}$, denoted by $\bm{w}$ for simplicity. 
The initialized real-valued parameters make up the over-parameterized neural network, while the agents aim to find their own subnetworks that is tailored to its specific statistical distribution and system requirements. 
To achieve this, each agent prunes its local personalized model by updating and sharing the binary mask tensor $\bm{m}_i$. 
{Compared with other methods based on LTH, such as those in \cite{li2020lotteryfl, dai2022dispfl} and as illustrated in Fig. \ref{sys2}(b), the key difference is whether real-valued parameters are transmitted or updated, which significantly reduces the communication cost in our approach.}

\subsection{Regularization and Binary Tensor}
Updating $\bm{z}_i$ directly according to \eqref{mask_update_z} may result in binary mask tensor with non-structured sparsity, and this irregular memory access can adversely impact practical acceleration in hardware implementations, leading to higher costs compared to structured sparsity. 
To resolve this issue, we incorporate a structured sparsity regularization term into the loss functions of the agents. 

Specifically, for the $l$-th convolutional layer, the real-valued mask tensor is denoted by $\bm{z}_{i,l}\in\mathbb{R}^{O_l\times I_l\times H_l\times W_l}$, where $O_l,I_l,H_l,W_l$ represent the dimensions of output filter, input channel, kernel height and kernel width. 
Group Lasso can effectively zero out all weights in certain groups, forming the basis of structured sparsity regularization.
In this work, we consider the sparsity along the output filter and take the entries in one output filter as a group.
Then the regularization term for the $l$-th convolutional layer in agent $i$ can be expressed as follows:
\begin{equation}
    \label{mask_z_regu_conv}
    R(\bm{z}_{i,l}) = \lambda\Big(\sum_{o_l=1}^{O_l}\|\bm{z}_{i,l}(o_l,:,:,:)\|_2\Big),
\end{equation}
where  $\lambda$ is the regularization parameter used to balance the loss and the pruning criterion.
Note that there may exist different forms of regularizers with different properties \cite{scardapane2017group, mitsuno2020hierarchical}, including $\ell_1$ norm, $\ell_2$ norm and composite $\ell_1/\ell_2$.
In this work, we do not focus on the specific regularizers. To simplify the computation of gradients, we use the $\ell_2$ norm regularization.   

Similarly, for the $l$-th linear layer, the regularization term for the real-valued mask tensor $\bm{z}_{i,l}\in\mathbb{R}^{O_l\times I_l}$ is as follows:
\begin{equation}
    \label{mask_z_regu_lin}
    R(\bm{z}_{i,l}) = \lambda\Big(\sum_{o_l=1}^{O_l}\|\bm{z}_{i,l}(o_l,:)\|_2\Big).
\end{equation}
By summing up \eqref{mask_z_regu_conv} and \eqref{mask_z_regu_lin} across all layers, we obtain the final regularization term as follows:
\begin{align}
    \label{mask_z_regu}
    R(\bm{z}_{i}) = & \lambda\sum_{l=1}^{L_c}\Big(\sum_{o_l=1}^{O_l}\|\bm{z}_{i,l}(o_l,:,:,:)\|_2\Big)\notag\\
    &+\lambda\sum_{l=1}^{L_l}\Big(\sum_{o_l=1}^{O_l}\|\bm{z}_{i,l}(o_l,:)\|_2\Big),
\end{align}
where $L_c$ and $L_l$ represent the number of convolutional and linear layers, respectively. 
Then, { the global loss function, with fixed global real-valued parameters $\bm{w}$ and updatable real-valued mask tensors $\bm{z}_i$ } can be expressed as follows:
\begin{equation}
    \label{opt_sys3}
    \min F(\textbf{Z}) := \frac{1}{N}\sum\nolimits_{i=1}^{N}f_i(\bm{w},\bm{z}_i),
\end{equation}
where $\textbf{Z}$ is the set of $\bm{z}_i$ for all node $i\in\mathcal{N}$. 
Here the local loss function of node $i$ is as follows:
\begin{equation}
    \label{opt_local}
    f_i(\bm{w},\bm{z}_i) := \mathbb{E}_{\mathcal{D}_i}[l_i(\bm{w},\bm{z}_i; \bm{x}_s, \bm{y}_s)]+R(\bm{z}_{i}).
\end{equation}

As discussed in Section \ref{sec2_b}, to obtain the binary mask tensor from the real-valued mask tensor, we use the threshold method \eqref{thres}. 
It is important to note that we still use the entry-wise ranking in each layer, instead of ranking the sum of absolute values in one filter and zeroing out those filters with smaller value. 
Although the latter method can ensure filter-wise sparsity after each iteration, it suffers from poor robustness and convergence properties. 
Therefore, we apply entry-wise ranking, along with the group sparsity regularization term, which regularizing the parameter tensor in each layer to the become structurally sparse. In this process, the hardware cost is reduced while guaranteeing the convergence performance.

Meanwhile, through experiments, we observe that the entry-wise ranking, combined with the structural sparsity regularization,  may lead to a small number of non-zero entries in some filters.
In such condition, these entries may have limited impact on the neural network accuracy while significantly increasing the hardware cost. 
To resolve this issue, we introduce an additional rule, denoted as $\text{Fil}[\cdot]$, which acts as follows: when the number of non-zero entries in one filter is below a threshold, all entries in the corresponding filter of the binary mask tensor are set to $0$. 
This operation can further enhance the structural sparsity of the local model. Then, for node $i$, the binary mask tensor in layer $l$ can be obtained from the real-valued tensor as follows:
\begin{equation}
    \bm{m}_{i,l}= \text{Fil}\big[\text{Thres}(\bm{z}_{i,l})\big].
\end{equation}

\subsection{Aggregation Algorithm Design}

\begin{figure*}[!htp]
    \centering
    \includegraphics[width=1.0\textwidth]{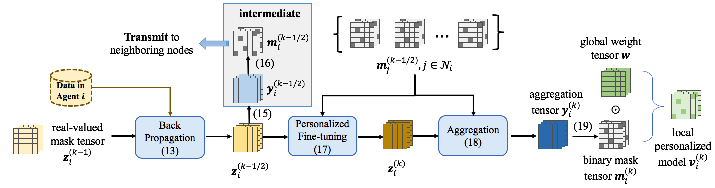}
    \caption{The algorithm flowchart of the proposed MCE-PL in the $k$-th iteration for node $i$: The updated and fused binary mask tensor $\bm{m}_i^{(k)}$ is obtained based on the aggregation tensor $\bm{y}_i^{(k)}$. $\bm{y}_i^{(k)}$ combines the information from $\bm{z}_i^{(k)}$ and the binary information $\bm{m}_j^{(k-1/2)}$ received from neighboring nodes $j\in\mathcal{N}_i$, where $\bm{z}_i^{(k)}$ is updated through back propagation and personalized fine-tuning.}
    \label{fig_algo}
    \vspace{-0.5cm}
\end{figure*}

The analysis in \cite{zhou2019deconstructing} shows the effectiveness of model learning using binary mask tensor. 
Experiments verify the hypothesis that the masking operation tends to direct toward the values they would have reached during training. Based on this observation, the entries in the mask tensor indicate whether the corresponding real-valued parameters are expected to approach 0. 
%Thus, the mask tensor contains certain information of the real-valued parameters. 
In distributed learning scenarios, although the data distribution among agents are non-i.i.d., the agents are located in some global environment and their personalized data distributions represent  partial views of the global data distribution.
Thus, certain correlations exist among agents, which can be leveraged through collaboration to enhance each model's performance. Since binary mask tensors contain specific information about the model parameters, they can be aggregated among agents to improve local model performance through appropriate algorithm design. 

The method in \cite{Li2021Fedmask} performs aggregation by taking intersection between the local updated mask and the other masks, and then initializing the real-valued mask tensor according to this result. 
However, this procedure is not suitable for heterogeneous system scenarios, where the retained model in each agent tends to converge to the smallest one, and the computational capacity in each agent cannot be sufficiently used. Thus, we need to carefully design the mask aggregation algorithm  that allows the fusion of information from neighboring nodes while preserving the heterogeneity of devices.
Given that the binary mask tensors received by each agent carry very limited information, it is challenging to exploit model relativity or to derive explicit aggregation weights as done in \cite{tian2023}. Therefore, we propose an alternative update framework based on a simple but effective method in personalized learning, which combines aggregation through averaging with an additional step of personalized fine-tuning. 

We first discuss the aggregation procedure.
Based on the observation from experiments, we find the key issue of the aggregation method: The direct aggregation of binary mask tensors on $\bm{z}_i$ consistently results in low accuracy and poor stability of the algorithm.  
This occurs because the direct operation on $\bm{z}_i$ cause the local model  to be heavily influenced by the neighboring binary mask tensors, whose limited information especially under non-i.i.d. condition leads to severe deviation of local model.  
To address this problem, in addition to the real-valued mask tensor $\bm{z}_i$, we introduce an another intermediate tensor $\bm{y}_i$ in each node, named as aggregation tensor.  
The aggregated binary mask in each agent is obtained based on this aggregation tensor. 

However, it is still challenging to fuse the binary information from $\bm{m}_j, j\in\mathcal{N}_i$ into the real-valued aggregation tensor.
As discussed above, in the binary mask tensor, the indices of those $1$ entries indicate larger amplitude of entries in the real-valued tensor. 
Based on this observation, a certain amplitude is expected to be added to the corresponding indices of the aggregated tensor. 
However, due to the limited information in the binary tensors, the explicit amplitude to be added could not be accurately computed. 
To address this challenge, we design an adaptive aggregation model to obtain the aggregation tensor $\bm{y}_i$ based on local $\bm{z}_i$ and the received binary mask tensors from neighboring nodes.

Considering the amplitude of $\bm{z}_i$ may vary among different agents, different layers in one agent and even across different iteration steps, we design an adaptively-defined value $\text{mean}[\text{abs}(\tilde{z}_{i,l})]$ to approximate the amplitude to be added for the $l$-th layer in node $i$.  Here the operation $\text{abs}(\cdot)$ means taking absolute values of all entries in the tensor and $\text{mean}[\cdot]$ take average of all entries in the tensor to get the mean value. The notation $\tilde{z}$ indicates that  the tensor is used for external operation and can be treated as a constant, which will not be involved in the gradient descent process. 
The aggregation is then performed by taking average of the received binary tensor. By considering the sign of $\bm{z}_i$, we derive the aggregation model of layer $l$ for agent $i$ as follows:
\begin{align}
    \label{mask_update1}
    \bm{y}_{i,l} = \bm{z}_{i,l} + \Big\{&\text{mean}\Big[\text{abs}\big(\tilde{\bm{z}}_{i,l}\big)\Big]\times\text{sign}(\tilde{\bm{z}}_{i,l})\notag \\ 
    &\odot\Big(\frac{1}{|\mathcal{N}_i|}\sum_{j\in\mathcal{N}_i}\bm{m}_{j,l}\Big)\Big\}. \end{align}

Based on the aggregation tensor, the corresponding aggregated binary mask tensor can be obtained through
\begin{equation}
    \label{mask_update8}
    \bm{m}_{i,l} = \text{Fil}\big[\text{Thres}(\bm{y}_{i,l})\big].
\end{equation}

On the other hand, the fine-tuning step with local data can adjust the direction of parameter updates to better align with the local data distribution. Inspired by this, we introduce a personalized fine-tuning step in each iteration to prevent model deviation caused by the limited information in mask tensors and the heterogeneous data distribution among agents.

The designed procedure is thus composed of three steps in each iteration: back propagation, personalized fine-tuning and model aggregation as shown in Fig. \ref{fig_algo}. 
In the following we describe the three steps one by one.

\textbf{Back Propagation.} Given the fixed real-valued parameters $\bm{w}$, together with the binary mask tensor updated in the last iteration $\bm{m}_{i}^{(k-1)}$, the forward computation can be conducted as follows:
\begin{equation}
    f_i(\bm{v}_i^{(k-1)}) = f_i(\bm{w}\odot\bm{m}_i^{(k-1)}).
\end{equation}

Based on the loss computed through the forward process, the gradient descent of $\bm{z}_i$ can be derived as follows:
\begin{align}
    \label{mask_update2}
        \bm{z}_i^{(k-1/2)} & = \bm{z}_i^{(k-1)}-\eta\frac{\nabla f_i}{\nabla \bm{v}_i}\times \frac{\nabla \bm{v}_i}{\nabla \bm{m}_i}\times \frac{\nabla \bm{m}_i}{\nabla \bm{z}_i}\notag \\
        & \approx \bm{z}_i^{(k-1)}-\eta\frac{\nabla f_i}{\nabla \bm{v}_i}\times \frac{\nabla \bm{v}_i}{\nabla \bm{m}_i}\times \text{sign}(\bm{z}_i^{(k-1)}),
\end{align}
where the gradient of $\bm{z}_i$  is given below,
\begin{equation}
    G(\bm{z}_i^{(k-1)}) = \frac{\nabla f_i}{\nabla \bm{v}_i}\times \frac{\nabla \bm{v}_i}{\nabla \bm{m}_i}\times \text{sign}(\bm{z}_i^{(k-1)}).
\end{equation}

According to \eqref{mask_update1}, the gradient of $\bm{z}_i$ equals to $\bm{y}_i$,  where the second term can be treated as an adaptive constant. 
Then,  based on $\bm{z}_i^{(k-1/2)}$, the intermediate aggregation tensor $\bm{y}_i$ and the intermediate binary tensor $\bm{m}_i$ can be obtained as follows:
\begin{subequations}    
\begin{align}
    \bm{y}_{i,l}^{(k-1/2)} =& \bm{z}_{i,l}^{(k-1/2)} + \Big\{\text{mean}\Big[\text{abs}\big(\tilde{\bm{z}}_{i,l}^{(k-1/2)}\big)\Big]\label{mask_update3} \\
    &\times\text{sign}(\tilde{\bm{z}}_{i,l}^{(k-1/2)})\notag \odot\Big(\frac{1}{|\mathcal{N}_i|}\sum_{j\in\mathcal{N}_i}\bm{m}_{j,l}^{(k-3/2)}\Big)\Big\}, \notag \\
% \end{align}
% \begin{align}
    \bm{m}_{i,l}^{(k-1/2)} =& \text{Fil}\big[\text{Thres}(\bm{y}_{i,l}^{(k-1/2)})\big].  \label{mask_update4}
    \end{align}
\end{subequations}
where the binary mask tensors from neighboring nodes $\bm{m}_{j,l}^{(k-3/2)}$ are those received in the former iteration.
Then each agent shares its intermediate binary mask tensor $\bm{m}_i^{(k-1/2)}$ to its neighboring nodes.

\textbf{Personalized Fine-tuning.} As mentioned before, to update the model in each agent towards the personalized data distribution, we add the fine-tuning step in each iteration. This step is essential especially in the mask-based aggregation, where the limited yet biased information from the binary tensors may lead to severe performance degradation. 
Specifically, considering the sparsity of those binary mask tensors, the aggregation makes an impact on some limited entries.
Thus, based on the received binary tensors $\bm{m}_j^{(k-1/2)}$, the fine-tuning is conducted on those entries which will be affected by the mask aggregation. 
Moreover, the approximated gradient of $\bm{z}_i$ obtained and stored before is utilized here to reduce the computational cost.
Then we can get the personalized fine-tuning step as follows:
\begin{equation}
    \label{mask_update5}
    \bm{z}_i^{(k)} = \bm{z}_i^{(k-1/2)} - \eta\times G(\bm{z}_i^{(k-1)})\odot\Big(\frac{1}{|\mathcal{N}_i|}\sum_{j\in\mathcal{N}_i}\bm{m}_j^{(k-1/2)}\Big).
\end{equation}
This fine-tuning step allows the model to aggregate in the subsequent step based on personalized needs, thereby enabling effective information fusion. 

\textbf{Aggregation.} Based on the fine-tuned $\bm{z}_i^{(k)}$, the aggregation of binary mask tensors in the aggregation tensor can be expressed by, 
\begin{align}
    \label{mask_update6}
    \bm{y}_{i,l}^{(k)} = \bm{z}_{i,l}^{(k)} + \Big\{&\text{mean}\Big[\text{abs}\big(\tilde{\bm{z}}_{i,l}^{(k)}\big)\Big]\times\text{sign}(\tilde{\bm{z}}_{i,l}^{(k)})\notag \\ 
    &\odot\Big(\frac{1}{|\mathcal{N}_i|}\sum_{j\in\mathcal{N}_i}\bm{m}_{j,l}^{(k-1/2)}\Big)\Big\}. \end{align}
Then, the aggregated binary tensor can be obtained as follows:
\begin{equation} \label{mask_update7}
    \bm{m}_{i,l}^{(k)} = \text{Fil}\big[\text{Thres}(\bm{y}_{i,l}^{(k)})\big].
\end{equation}
Note that,  the aggregation step \eqref{mask_update7} is conducted after the personalized fine-tuning, which is different from the typical aggregation-and-fine-tuning procedure. This is necessary because the introduced aggregation tensor needs to be derived based on the fine-tuned $\bm{z}_i$. 
According to the above three steps, the Mask-based Communication-Efficient Personalized Learning Algorithm (MCE-PL) can be summarized in Algorithm \ref{Alg1}.

\begin{algorithm}[!t]
	\caption{\textbf{MCE-PL}}\label{Alg1}
	\For{node $i=1,2,...,N$}
	{
		\textbf{Initialize} the real-valued parameters $\bm{w}$ and binary mask tensor from neighboring nodes $\bm{m}_j^{(0)}, j\in\mathcal{N}_i$. \textbf{Set} $k=0$.\\
	}
	\While{not converge}
	{
		$k=k+1$\\
		\For{node $i=1,2,...N$}
		{
			\textbf{Randomly select} a batch of data from local training data set $\{\bm{x}_s, \bm{y}_s\}\in\mathcal{D}_i$.\\
			\textbf{Compute} the loss by (\ref{opt_local}).\\
			\textbf{Update} $\bm{z}_i^{(k-1/2)}$ by back propagation and get $G(\bm{z}_i^{(k-1)})$.\\
			\textbf{Obtain} $\bm{y}_i^{(k-1/2)}$ and $\bm{m}_i^{(k-1/2)}$ by (\ref{mask_update3}) and (\ref{mask_update4}).\\
		        \textbf{Transmit} $\bm{m}_{i}^{(k-1/2)}$ to all neighboring nodes. %$j\in\mathcal{N}_i$.
		}
		\For{node $i=1,2,...N$}
		{
			\textbf{Personalized fine-tune} and get the updated $\bm{z}_i^{(k)}$ by (\ref{mask_update5}).\\
			\textbf{Aggregate} the information and get the aggregated binary tensor $\bm{m}_i^{(k)}$ by (\ref{mask_update7}). \\
		}
	}
	\textbf{Output} the personalized model in each agent.
\end{algorithm}
Note that, each agent does not need to keep the intermediate tensors $\bm{y}_i^{(k-1/2)}$ and $\bm{y}_i^{(k)}$ after obtaining $\bm{m}_i^{(k-1/2)}$ and $\bm{m}_i^{(k)}$. 
Likewise, $\bm{m}_i^{(k-1/2)}$ can be relaxed after computing $\bm{z}_i^{(k-1/2)}$. 
% Thus, compared to the traditional update based on binary tensor, although MCE-PL introduces some intermediate tensors, its storage cost only adds the part of approximated gradient $G(\bm{z}_i^{(k-1)})$. 

\newtheorem{remark}{Remark}
{
\begin{remark}
\textbf{Computational Complexity.} 
The computational complexity has been widely investigated in the context of real-value-based updating, with the derived convergence rate. For distributed stochastic gradient descent (DSGD) \cite{Lian2017}, its convergence rate is $\mathcal{O}(\frac{1}{\sqrt{NT}})$, and the computational complexity can be derived as $\mathcal{O}(\frac{1}{\epsilon^2})$ concerning the $\epsilon$-approximation solution. 
However, the convergence analysis for the binary-mask-based updating is still a challenging topic, even in centralized settings \cite{frankle2018lottery}. 
So here we simplify the computational complexity analysis in one iteration.
In each iteration, the (intermediate) aggregation tensor $\bm{y}_i$ and the (intermediate) binary tensor $\bm{m}_i$ are updated in a layer-wise way within each agent. 
If we further take the aggregation in (\ref{mask_update3}) and (\ref{mask_update6}) into consideration, under large $N$, the total number of neighboring nodes can be approximated by $\mathcal{O}(N^2p)$ with connectivity probability $p$. 
Thus, given $N$ agents, the computational complexity over the network in one iteration is $\mathcal{O}(N^2Lp)$ considering the neural network with $L$ layers.
Compared with FedProx \cite{li2020federated}, the computation cost w.r.t. time and energy of the proposed algorithm may increase with the additional pruning steps defined by (\ref{mask_update4}) and (\ref{mask_update7}). Compared with LotteryFL \cite{li2020lotteryfl}, the computation cost of MCE-PL is lightly higher with the personalized fine-tuning step.
\end{remark}
}

{
\begin{remark}
\textbf{Privacy and Security Analysis.}
Note that our approach leverages a unique parameterization in which each local model is obtained by the Hadamard product of fixed global parameters and personalized binary masks. This configuration not only introduces structural obfuscation—effectively discretizing the parameter space and introducing quantization noise into the local model—but also achieves information fragmentation by restricting updates solely to the binary masks while keeping the global parameters constant. Consequently, the temporal correlation between successive updates is disrupted, significantly mitigating the risks of model inversion and information leakage common in conventional gradient-based methods.
\end{remark}
}

\section{Theoretical Proof of DSLTH}\label{Sec4}
In this section, we give the theoretical proof of the proposed distributed strong lottery ticket hypothesis under the statistical heterogeneous conditions, based on the rigorous proof of SLTH in \cite{da2022proving}.
For the sake of simplicity, we consider the convolutional neural network in a restricted setting, which is defined as $f: [0,1]^{D\times D\times O_0} \to \mathbb{R}^{D\times D\times O_L}$ with the form
\begin{equation}
	\label{form_f}
	f(\bm{x}) = \bm{v}_L\ast\sigma(\bm{v}_{L-1}\ast\cdots\sigma(\bm{v}_{1}\ast\bm{x})).
\end{equation}
Here, $\ast$ denotes the convolution operation, $\sigma(\cdot)$ represents the ReLU activation function and $\bm{v}_l\in\mathbb{R}^{o_l\times o_{l-1}\times d_l\times d_l}$ represents the convolution weight in the $l$-th layer, where the convolutions have no bias and are suitably padded with zeros. 
Note $f$ does not include additional operations such as stride or average pooling, a more general analysis is deferred to future works. 

In the distributed scenarios with heterogeneous data, we consider an arbitrary pair of nodes, whose relationship can be extended to those nodes in the whole decentralized network without loss of generality.
Use the subscript $1$ and $2$ to denote the arbitrary two nodes, and their local datasets are denoted by $\mathcal{D}_1=\{\bm{x}_{1,s}, \bm{y}_{1,s}\}$ and $\mathcal{D}_2=\{\bm{x}_{2,s}, \bm{y}_{2,s}\}$, including both training and testing data due to the personalized problem. 
The two local datasets have personal distribution over some common feature space $\mathcal{X}$ and label space $\mathcal{Y}$, with the dimension ${D\times D\times I_0}$ of $\bm{x}, \forall \bm{x}\in\mathcal{X}$.
We define the two local models as $f_1$ and $f_2$ for any two nodes, which are sufficiently and independently trained by local dataset from the same initialization $f_0$ with the architecture/form in (\ref{form_f}).

For a tensor $\bm{a}$, the notation $\|\bm{a}\|_1$ is its $\ell_1$ norm defined as the sum of the absolute values of each entry in $\bm{a}$. 
$\|\bm{a}\|$ is the $\ell_2$ norm, where $\|\bm{a}\|^2$ equals to the sum of the each entry's square value. 
And the notation $\|\bm{a}\|_{\max}$ refers to its maximum norm as the maximum among the absolute value of each entry.
Define a class of functions from ${[0,1]}^{D\times D\times O_0}$ to $\mathcal{R}^{D\times D\times O_l}$ as $\mathcal{F}$ such that for each $f\in\mathcal{F}$,
$f(\bm{x}) = \bm{v}_L\ast\sigma(\bm{v}_{L-1}\ast\cdots\sigma(\bm{v}_{1}\ast\bm{x}))$, where for each $l\in[L]$, the weight tensor $\bm{v}_l\in{[-1, 1]}^{O_l\times O_{l-1}\times d_l\times d_l}$ and $\|\bm{v}_l\|_1\le1$.

We firstly make some assumptions for simplicity of analysis.
 
\newtheorem{assumption}{Assumption}
\begin{assumption}
\label{as1}
In the dataset, each entry of the features is in the range of $[0,1]$, i.e., $\bm{x}\in{[0, 1]}^{D\times D\times I_0}, \forall \bm{x}\in\mathcal{X}$.
\end{assumption}

\begin{assumption}
\label{as2}
The two sufficiently and independently trained networks $f_1, f_2\in\mathcal{F}$.
\end{assumption}

\begin{assumption}
\label{as4}
The output distance of $f_1$ and $f_2$ can be upper bounded by $\sup_{\bm{x}\in\mathcal{X}}\|f_1(\bm{x})-f_2(\bm{x})\|_{\max}\le\alpha_u$, with the different parameters indicated by $\inf_{\bm{x}\in\mathcal{X}}\|f_1(\bm{x})-f_2(\bm{x})\|_{\max}\ge\alpha_l$.
\end{assumption}

Among these assumptions, Assumption \ref{as1} can be satisfied through data normalization. Assumption \ref{as2} can be satisfied through the proper initialization and training process such that there exist $f_1$ and $f_2$ in the class of $\mathcal{F}$.
The upper bound in Assumption \ref{as4} measures the distance between the output of two models given the same input. It is proposed and rational considering the same initialization $f_0$ and the correlation between the local datasets in the two nodes, which are located in a global data distribution. It should be noted that the value $\alpha_u$ would be related to the distance between the parameters of $f_1$ and $f_2$, which can be further correlated to distance between gradients in the training process.
We defer to future works for a more comprehensive investigation in such non i.i.d. conditions.
On the other hand, considering the personalized data distribution, the difference of the learned models can be indicated by $\inf_{\bm{x}\in\mathcal{X}}\|f_1(\bm{x})-f_2(\bm{x})\|_{\max}\ge\alpha_l$, meaning that for any data sample, there exists certain distance between the output, which is rational due to the independent training in the non-i.i.d. condition.
Then the theory on SLTH in centralized condition is shown as follows:

\newtheorem{lemma}{Lemma}
\begin{lemma}\label{theo1}
(\textit{Theorem 1 in \cite{da2022proving}} for SLTH) Consider a convolutional network with $L$ layers. $\varepsilon$ and $C$ are two positive constant. Let $O_l, I_l\in\mathbb{N}$ with $I_l\ge CO_l\log\frac{O_{l-1}O_ld_l^2L}{\min\{\varepsilon, \delta\}}$. Define $\bm{w}_{2l-1}\in\mathbb{R}^{I_l\times O_{l-1}\times d_i\times d_i}$, $\bm{w}_{2l}\in\mathbb{R}^{O_l\times I_l\times 1\times 1}$. Define the corresponding binary mask tensors with the same shape, i.e., $\bm{m}_{2l-1}\in\{0,1\}^{\text{size}(\bm{w}_{2l-1})}$ and $\bm{m}_{2l}\in\{0,1\}^{\text{size}(\bm{w}_{2l})}$. The entries of $\bm{w}_1,...,\bm{w}_{2L}$ are i.i.d. random variables from the uniform distribution in $[-1, 1]$. Define the random $2L$-layer CNN $g: {[0,1]}^{D\times D\times O_0}\to\mathcal{R}^{D\times D\times O_l}$ as: 
\begin{equation}
\label{form_g}
g(\bm{x})=\bm{w}_{2L}\ast\sigma(\cdots\sigma(\bm{w}_{1}\ast \bm{x})).
\end{equation}
Its pruned version is defined as:
\begin{equation*}
g_{\bm{m}}(\bm{x})=(\bm{w}_{2L}\odot\bm{m}_{2L})\ast\sigma[\cdots\sigma[(\bm{w}_{1}\odot\bm{m}_1)\ast \bm{x}].
\end{equation*}
Then we can choose constant $C$ independently from other parameters so that with probability at least $1-\delta$, the following holds for every $f\in\mathcal{F}$:
\begin{equation}
\label{eq_theo1}
\inf_{\bm{m}\in\{0,1\}^{\text{size}(\bm{w})}}\sup_{\bm{x}\in{[0,1]}^{D\times D\times O_0}}\big\|f(\bm{x})-g_{\bm{m}}(\bm{x})\big\|_{\max}\le\varepsilon.
\end{equation}
\end{lemma}

In \eqref{eq_theo1} of Lemma \ref{theo1}, $\inf$ indicates the existence of such pruned network.
Then based on Lemma \ref{theo1} and the definitions in it, in the distributed condition, we can give the following major theorem for an arbitrary pair of nodes denoted by subscript $1$ and $2$ as talked above:

\newtheorem{theorem}{Theorem}
\begin{theorem}\label{theo2}(DSLTH)
Suppose the trained model $f_1, f_2\in\mathcal{F}$. Define the random 2L-layer CNN $g$ as that in (\ref{form_g}). Its pruned versions in node $1$ and node $2$ are respectively defined as:
\begin{equation*}
g_{\bm{m}_1}(\bm{x})=(\bm{w}_{2L}\odot\bm{m}_{1,2L})\ast\sigma[\cdots\sigma[(\bm{w}_{1}\odot\bm{m}_{1,1})\ast \bm{x}],
\end{equation*}
\begin{equation*}
g_{\bm{m}_2}(\bm{x})=(\bm{w}_{2L}\odot\bm{m}_{2,2L})\ast\sigma[\cdots\sigma[(\bm{w}_{1}\odot\bm{m}_{2,1})\ast \bm{x}].
\end{equation*}
Then there exist binary mask tensors $\bm{m}_{1}$ and $\bm{m}_{2}$, such that the pruned networks in node $1$ and $2$ have the following property with probability at least ${(1-\delta)}^2$:
\begin{equation} 
\label{eq3_th2}
\sup_{\bm{x}\in\mathcal{X}}\big\|f_1(\bm{x})-g_{\bm{m}_1}(\bm{x})\big\|_{\max}\le\varepsilon_1, %\inf_{\bm{m}_{1}\in\{0,1\}^{\text{size}(\bm{w})}} {[0,1]}^{D\times D\times O_0}
\end{equation}
\begin{equation}
\label{eq4_th2}
\sup_{\bm{x}\in\mathcal{X}}\big\|f_2(\bm{x})-g_{\bm{m}_2}(\bm{x})\big\|_{\max}\le\varepsilon_2. %\inf_{\bm{m}_{2}\in\{0,1\}^{\text{size}(\bm{w})}}
\end{equation}
Moreover, the pruned networks between node $1$ and $2$ has the following correlation with probability ${(1-\delta)}^2$: 
\begin{equation}
\label{eq1_th2}
\sup_{\bm{x}\in\mathcal{X}}\big\|g_{\bm{m}_1}(\bm{x})-g_{\bm{m}_2}(\bm{x})\big\|_{\max}\le\varepsilon_1+\varepsilon_2+\alpha_u, %\inf_{\bm{m}_{1}, \bm{m}_{2}\in\{0,1\}^{\text{size}(\bm{w})}}
\end{equation}
and the heterogeneity with probability ${(1-\delta)}^2$:
\begin{align}
\label{eq2_th2}
&\inf_{\bm{x}\in\mathcal{X}}\big\|g_{\bm{m}_1}(\bm{x})-g_{\bm{m}_2}(\bm{x})\big\|_{\max} \notag \\
& \ge \min\{|\varepsilon_1+ \varepsilon_2-\alpha_l|, |\varepsilon_1+ \varepsilon_2-\alpha_u|, |\alpha_l|\}.
\end{align}
\end{theorem}
\noindent \textit{Proof:}  See Appendix~\ref{appendix:proof Theorem1}. \hfill $\blacksquare$

{Theorem \ref{theo2} establishes the theoretical foundation for DSLTH and the proposed framework. \eqref{eq3_th2} and \eqref{eq4_th2} demonstrate the good performance of the locally pruned subnetworks.
Additionally, under the same real-valued parameters in $g_{\bm{m}_1}(x)$ and $g_{\bm{m}_2}(x)$, \eqref{eq1_th2} reveals the correlation between the binary mask tensors $\bm{m_1}$ and $\bm{m}_2$, illustrating the effectiveness of inter-agent  correlation. Meanwhile, \eqref{eq2_th2} indicates their heterogeneity and the necessity of personalization.}

It should be noted that Theorem \ref{theo2} only verifies DSLTH in heterogeneous data distribution, without showing the impact of system heterogeneity. 
Moreover, Theorem \ref{theo2}, following Lemma \ref{theo1}, only focuses on the convolutional layers with the specified form of $g$, which is not general to practical deep neural networks that combine other type of layers.
To further investigate the proposed DSLTH, we refer to simulations presented in Section~\ref{veri_DSLTH}.

\section{Numerical Experiments}\label{experiment}
In this section, we first experimentally verify the proposed distributed strong lottery ticket hypothesis, to show its effectiveness in data and node heterogeneity. Then the performance of the proposed MCE-PL is simulated and compared with other methods, to show its improved accuracy, faster convergence rate and communication-efficiency under the heterogeneous scenario. Consider the $10$-class classification problem in CIFAR-10 dataset and apply the widely-used AlexNet architecture in each agent, which is made of three $5\times 5$ convolutional layers, each followed by a $3\times 3$ max pooling layer with stride 2, and two fully connected layers.

\subsection{Verification of Distributed Strong Lottery Ticket Hypothesis}\label{veri_DSLTH}
\begin{figure*}[!htp]
    \centering
    \includegraphics[width=0.9\textwidth]{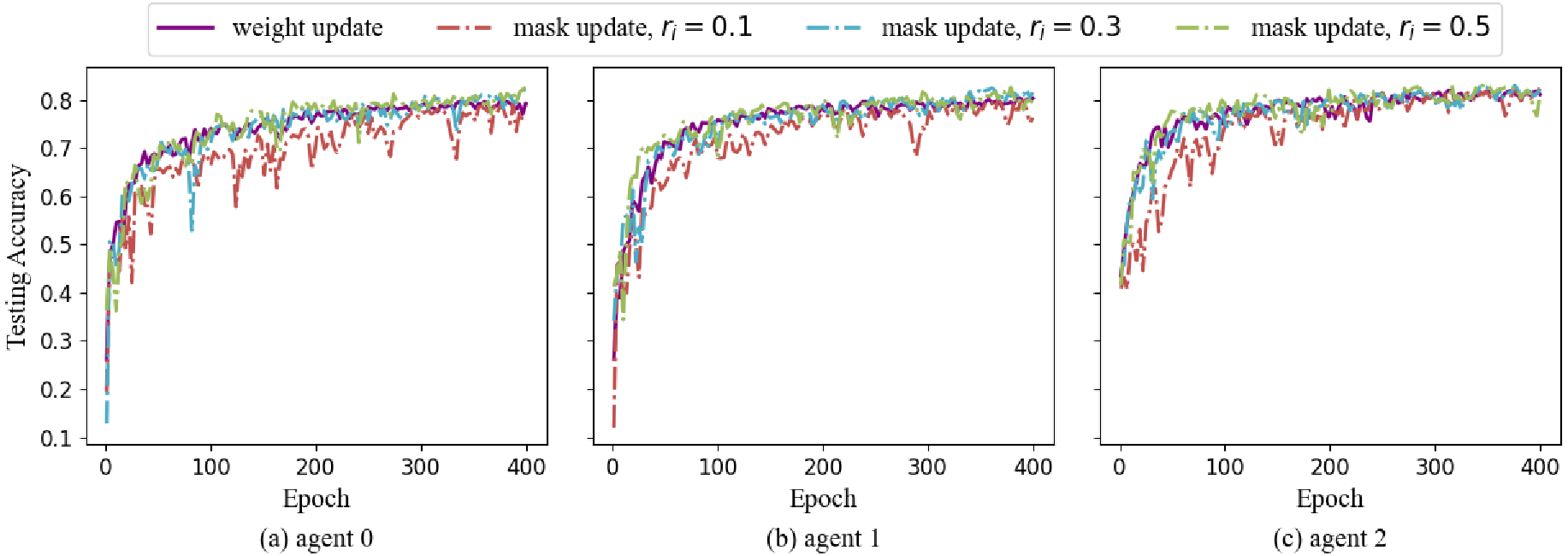}
    \caption{The verification of the DSLTH: comparison of convergence curves between the weight-based update and the mask-based one considering $r_i=0.1, 0.3, 0.5$ in different agents.}
    \label{exp2}
    \vspace{-0.5cm}
\end{figure*}

In addition to the analysis in Section \ref{sec3_a} and theoretical proof in Section \ref{Sec4}, in this subsection we verify the proposed DSLTH through simulations. Specifically, we choose three agents, each with three different labels in CIFAR-10 dataset. Note here we mainly aim to validate the performance of the pruned subnetworks, and the analytical results in \ref{Sec4} cannot be practically applied here.
Thus, different from the construction of $f$ and $g$ in Section \ref{Sec4}, in this part we simulate a more general neural network. 
We want to emphasize that the experimental results aim to show the learning ability and accuracy of the pruned subnetworks.
Considering the complicated construction of $g$ in Section \ref{Sec4}, we choose the network architecture of $g$ the same as the baseline architecture, and is not strictly required to have twice the number of layers. 

Here, in order to show the fully-trained results, each agent has all of the data samples corresponding to its labels. 
The parameters are initialized randomly from the uniform distribution in $[-1,1]$.
Firstly, the parameters of each model are sufficiently trained with the local heterogeneous dataset, which is named as weight update.
Then based on the same and fixed initialized parameters, each local model is trained with binary mask update procedure as described in Section \ref{sec2_b}, to obtain the pruned subnetworks of the original neural network. 
We set $r_i=0.1, 0.3, 0.5$ respectively to show that some well-performed subnetworks could be found even under different $r_i$. 
The accuracy is tested every $3$ update steps and the results are shown in Fig. \ref{exp2}. 

It can be observed that the pruned subnetworks on the randomly initialized network can achieve similar accuracy as the fully-trained network, both in different agents (data heterogeneity) and different $r_i$ (system heterogeneity). The convergence curves of the update based on mask show more oscillation compared with weight update. This is because the update based on mask is inherently a discrete update, where the updated subnetworks between iterations may lead to divergent accuracy. The above results verify the proposed DSLTH in a simple yet efficient way. Based on these fundamental results, in the following we simulate on the proposed method MCE-PL and compare it with other methods.

\subsection{Comparison on Convergence and Communication Cost in Data Heterogeneity}
{In this subsection, we first compare the convergence results and communication cost of the proposed methods with other state-of-the-art algorithms considering only statistical heterogeneity.
{Specifically, we compare the proposed MCE-PL with the conventional decentralized stochastic gradient descent (DSGD) and FL. We also implement the FedProx \cite{li2020federated} and LotteryFL \cite{li2020lotteryfl} algorithms in a decentralized network, referring to them as DSGDProx and LotteryDSGD, respectively. Additionally, we compare the standalone method, where each agent performs model updates and pruning independently, as described in Section \ref{sec2_b}, referred to as ind-mask.}

We consider a multi-agent communication network with $N=20$ nodes, whose topology is generated randomly using the \textit{Erdos\_Renyi} graph model, with the connectivity probability equal to $p=0.5$.  
%The randomly generated communication topology is shown in Fig.
Consider the $10$-class classification problem in CIFAR-10 dataset, where we randomly choose $c_i$ labels assigned for each agent to reflect the non-i.i.d. data distribution. 
The training samples corresponding to the same label are averaged and randomly assigned to the agents, while the testing samples corresponding to local labels are all assigned to each agent for a general and comprehensive evaluation of the learned model. 

We simulate the performance of algorithms under different learning rate $[1.0, 0.1, 0.01, 0.001]$ and choose the learning rate that yields the best performance. Specifically, the learning rate is set to $1.0$ for the binary mask-based algorithm MCE-PL and ind-mask and $0.1$ for LotteryDSGD. For the other algorithms based on real-valued parameters, the learning rate is set to $0.01$. 
Notably, the learning rate for $\bm{z}_i$ is much larger than that for $\bm{w}_i$. This is due to the update mechanism for $\bm{z}_i$, where the binary mask tensor is updated based on the sorted absolute values of its entries. Therefore, a larger learning rate induces significant changes in the amplitude of $\bm{z}_i$, affecting the sorting results and, consequently, the binary mask tensor. The regularization term parameter is set to $\lambda = 0.001$. The results are presented in Table \ref{exp7}, with sparsity requirements $0.3$ and $0.5$.

\begin{table}[!htp]
    \centering
    \small
    \caption{The comparison of averaged testing accuracy among different algorithms under data heterogeneity.}
    \label{exp7}
    \small
    \setlength{\tabcolsep}{4.5mm}{
    \begin{tabular}{l|l|l|l}
    \hline
    & \multicolumn{1}{l|}{$c_i=3$} & \multicolumn{1}{l|}{$c_i=4$} & \multicolumn{1}{l}{$c_i=5$} \\ \hline
    DSGD & 0.6218     & 0.5811      & 0.5660 \\
    % \hline
    FL & 0.6112    & 0.6097    & 0.6072  \\
    % \hline
    DSGDProx & 0.7550    & 0.6863    & 0.6143  \\
    % \hline
    LotteryDSGD (0.3) & 0.6687    & 0.6212    & 0.5755  \\
    % \hline
    LotteryDSGD (0.5) & 0.5612    & 0.6686    & 0.5217  \\
    % \hline
    ind-mask (0.3) & 0.7092    & 0.6520    & 0.5876  \\
    % \hline
    ind-mask (0.5) & 0.6934    & 0.6740    & 0.5922  \\
    \hline
    MCE-PL (0.3) & 0.7498    & \textbf{0.6951}    & 0.6048  \\
    % \hline
    MCE-PL (0.5) & \textbf{0.7603}    & 0.6943    & \textbf{0.6244}  \\
    \hline
    \end{tabular}}
    %\vspace{-0.5cm}
\end{table}

As shown in Table \ref{exp7}, the proposed MCE-PL achieves good performance in terms of convergence accuracy. Regarding communication cost, we compare the decentralized protocols under $c_i=3$, focusing on LotteryDSGD with a $0.3$ sparsity ratio and MCE-PL with a $0.5$ sparsity ratio, which demonstrate better performance. The convergence curves, in relation to communication cost, are displayed in Fig. \ref{exp8}, where the proposed MCE-PL achieves the highest testing accuracy with significantly lower communication cost.

\begin{figure}[!htp]
    \centering
    \includegraphics[width=0.4\textwidth]{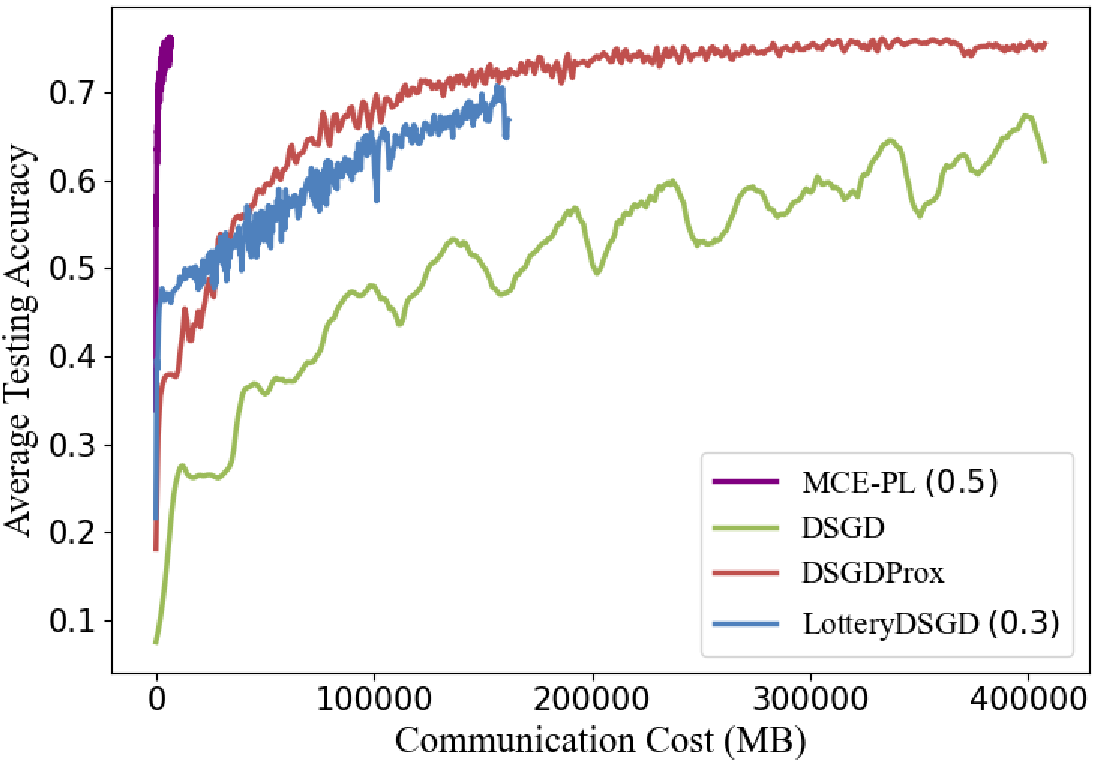}
    \caption{The comparison of communication cost.}
    \label{exp8}
    \vspace{-0.5cm}
\end{figure}
}

\subsection{Comparison on Convergence and Communication Cost in Node Heterogeneity}
{All of the above baseline methods cannot be applied in node heterogeneity settings. To further show the superiority of the proposed method, we additionally consider the node heterogeneity in this subsection.} 
To indicate the different communication, storage and computation capabilities among agents, they have varying model retention ratios $r_i$. 
It needs to be noted that the optimization of the retention ratio among agents considering load balancing is another interesting issue, which however is not the focus of this work. So here in the simulation, we randomly generate different $r_i$ in the range of $[0.1, 0.2, 0.3, 0.4]$ for a general test of the algorithm's performance, and the generated sparsity requirements of agents are $0.4, 0.4, 0.3, 0.2, 0.1, 0.1, 0.1, 0.2, 0.4, 0.2, 0.4, 0.2, 0.3, 0.3, 0.4,$ $0.3, 0.1, 0.1, 0.3, 0.3$ respectively.

We compare the proposed MCE-PL with several different algorithms {in decentralized setting applying typical techniques and suited for the node heterogeneity}.
\begin{itemize}
\item{\textbf{ind-mask}: Each agent conducts model update and model pruning independently based on the method in Section \ref{sec2_b}, without collaboration and information fusion.}
\item{\textbf{ind-weipru}: Each agent conducts model update with conventional gradient descent on real-valued parameters. After each iteration, each agent prunes its local model according to absolute values under the requirements of model retention ratio.}
\item{\textbf{avr-weipru}: Each agent updates its local model with conventional gradient descent on real-valued parameters, then prunes the model and transmits it to neighboring nodes. The agents then aggregate the models by taking average of the received model parameters.}
\item{\textbf{par-weipru}: Each agent updates its local model with conventional gradient descent on real-valued parameters, then prunes the model and transmits it to neighboring nodes. The agents then aggregate the models by partially averaging.}
\end{itemize}

\begin{figure}[!htp]
    \centering
    \includegraphics[width=0.44\textwidth]{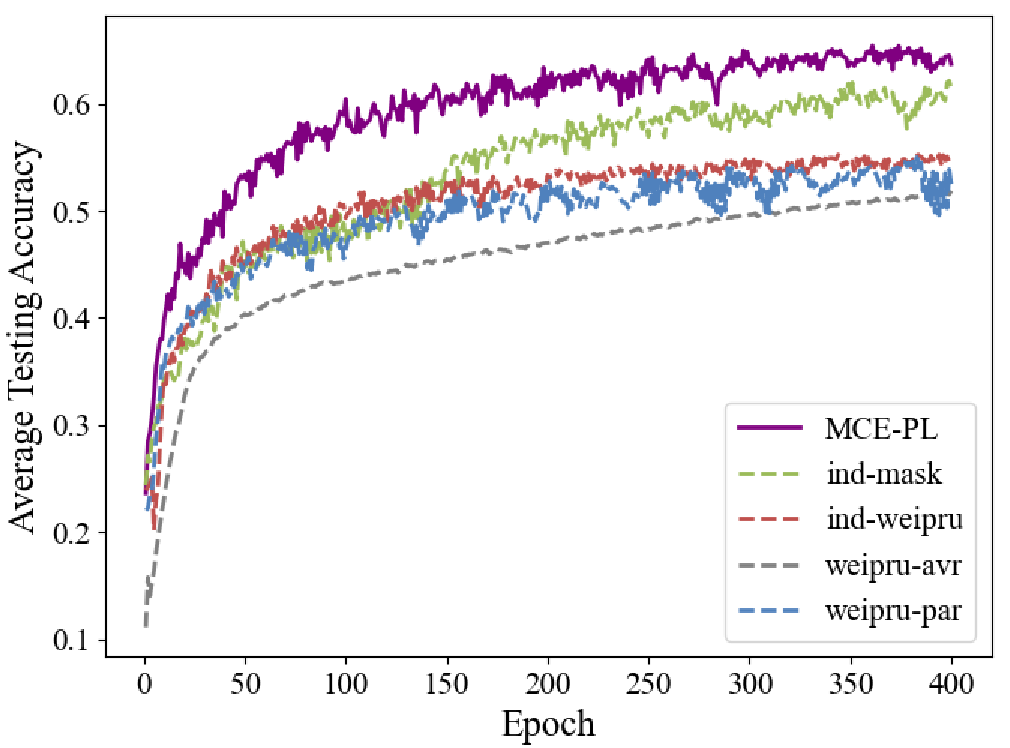}
    \caption{The convergence curves of different algorithms.}
    \label{exp1}
    % \vspace{-0.5cm}
\end{figure}

In each iteration, the gradient is computed with stochastic gradient descent, where the batch size is 128. The learning rates are also selected from $[1.0, 0.1, 0.01, 0.001]$ and choose those with best performance. The learning rate is set to $1.0$ for the binary mask-based algorithm MCE-PL and \textbf{ind-mask} and $0.001$ for the other algorithms. The number of labels in each agent is set to $c_i=4$, where the labels are assigned to each agent randomly. The convergence curves of the algorithms are shown in Fig. \ref{exp1}.

Fig. \ref{exp1} compares the convergence performance of different algorithms. It could be observed that the oscillation exists on most of the curves, owing to the model pruning in each iteration, while the procedure \textbf{avr-weipru} reduces the divergence by taking the average. 
We then compare the two algorithms based on binary mask tensor, MCE-PL and \textbf{ind-mask}. It can be seen that compared to independent update, the cooperation among agents can fasten the convergence rate and improve the model accuracy, showing the effectiveness of the designed algorithm. Then we compare the two independent methods \textbf{ind-mask} and \textbf{ind-weipru}. In the early iterations, \textbf{ind-mask} has slower convergence rate, while its eventual accuracy is higher than \textbf{ind-weipru}. It is resulted from the model pruning in \textbf{ind-weipru}. Finally, the partially averaging method \textbf{par-weipru} performs better than \textbf{avr-weipru}, due to the heterogeneity among agents. However, it performs worse than \textbf{ind-weipru}, showing its limited effectiveness in such highly heterogeneous scenarios. 
Compared to these baseline algorithms, the proposed MCE-PL has the fastest convergence rate and the highest test accuracy.

{We also compared the proposed MCE-PL with Dis-PFL \cite{dai2022dispfl} to evaluate its performance.}
The model testing accuracy of these algorithms under different number of local labels $c_i=3,4,5$, as shown in Table. \ref{exp6}.
It can be seen that the proposed MCE-PL outperforms the other baseline methods in this label distribution skew condition. With the increase of $c_i$, the advantages of MCE-PL become more significant and the cooperation can be more effective, due to the enhanced relativities of data distributions among agents.

\begin{table}[!htp]
    \centering
    \small
    \caption{The comparison of averaged testing accuracy among different algorithms {under data and node heterogeneity.}}
    \label{exp6}
    \small
    \setlength{\tabcolsep}{5mm}{
    \begin{tabular}{l|l|l|l}
    \hline
    & \multicolumn{1}{l|}{$c_i=3$} & \multicolumn{1}{l|}{$c_i=4$} & \multicolumn{1}{l}{$c_i=5$} \\ \hline
    MCE-PL & \textbf{0.7088}     & \textbf{0.6451}      & \textbf{0.5759} \\
    \hline
    ind-mask & 0.6808    & 0.6140    & 0.5413  \\
    \hline
    ind-weipru & 0.6239    & 0.5484    & 0.4813  \\
    \hline
    weipru-avr & 0.5515    & 0.5187    & 0.4564  \\
    \hline
    weipru-par & 0.6068    & 0.5251    & 0.4643  \\
    \hline
    {Dis-PFL} & {0.5742}    & {0.5609}    & {0.5617}  \\
    \hline
    \end{tabular}}
\end{table}

\begin{figure}[!htp]
    \centering
    \includegraphics[width=0.44\textwidth]{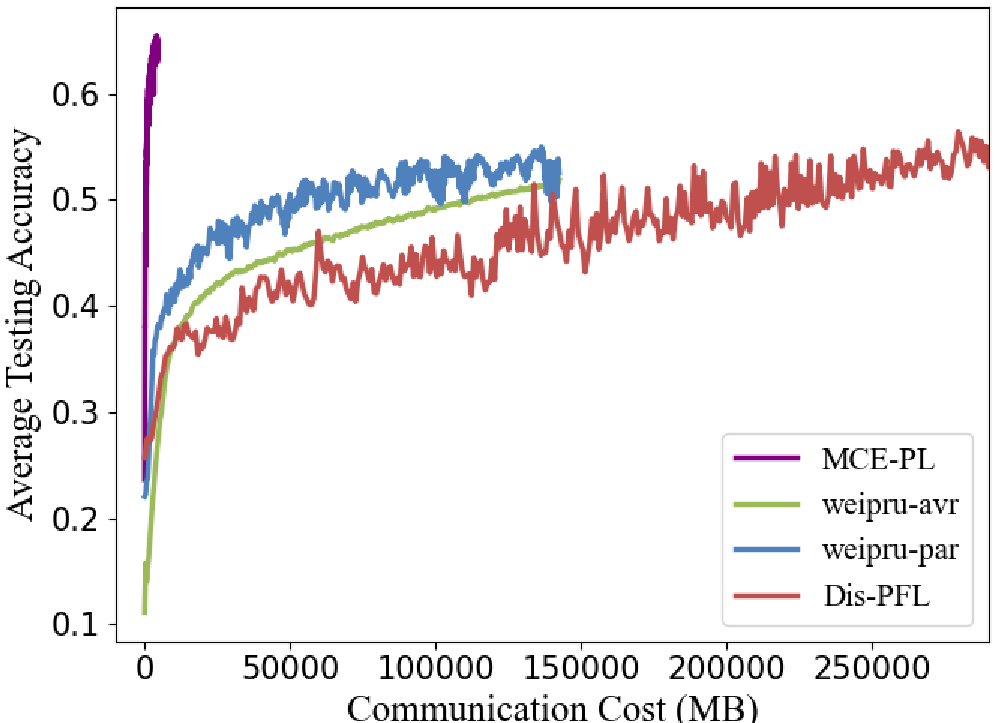}
    \caption{The comparison of communication cost.}
    \label{exp5}
    \vspace{-0.2cm}
\end{figure}

Finally, we show the communication cost curves of the collaborative methods, MCE-PL, \textbf{avr-weipru} and \textbf{par-weipru}. Under the same simulation settings as above with $c_i=4$, the communication cost is measured by the amount of data transmitted. 
For \textbf{avr-weipru} and \textbf{par-weipru} based on real-valued parameters, the size of each transmission package is equal to the number of the parameters, where each data is a 32 bit float. For MCE-PL, the size of each binary data is 1 bit. Then we get the curves of test accuracy with the communication cost in Fig. \ref{exp5}.

It can be observed from Fig. \ref{exp5} that the proposed MCE-PL has much less communication cost compared with weipru-avr and weipru-par based on real-valued parameters. 
{Additionally, Dis-PFL has higher communication costs due to the transmission of both the pruned model and binary masks, while the mask-growing technique in Dis-DFL results in a denser model than the intended sparsity. Furthermore, MCE-PL achieves higher accuracy at the same communication cost, demonstrating its communication efficiency.}

\subsection{Comparison of Learned Model}
\begin{figure*}[!htp]
    \centering
    \includegraphics[width=0.8\textwidth]{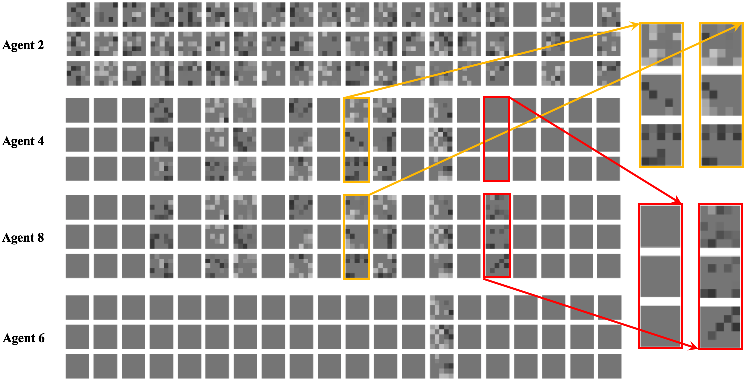}
    \caption{The learned personalized models in different agents by MCE-PL.}
    \label{exp4_a}
    \vspace{-0.2cm}
\end{figure*}

\begin{figure*}[!htp]
    \centering
    \includegraphics[width=0.8\textwidth]{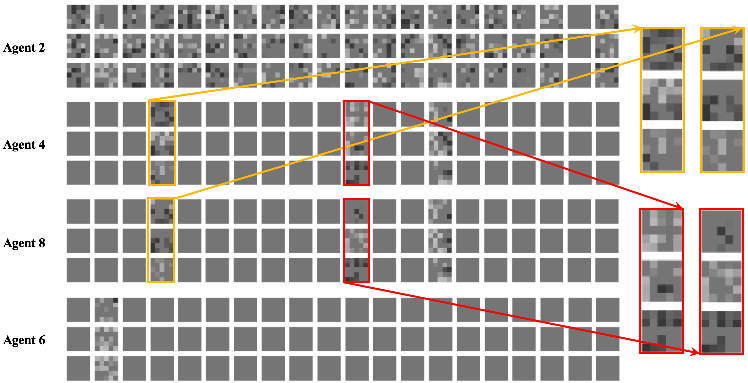}
    \caption{The learned personalized models in different agents by \textbf{ind-mask}.}
    \label{exp4_b}
    \vspace{-0.5cm}
\end{figure*}

Then given the above simulation settings, following we show the comparison of the learned models in MCE-PL and \textbf{ind-mask}. Specifically, consider node $2,4,6,8$ whose model retention ratios are $0.4, 0.2, 0.1, 0.2$ respectively. For the first convolutional layer with size $3\times 64\times5\times5$ in each agent, we plot part of the learned model, to display the results more clearly and concisely. We select the middle $20$ filters and the convolutional kernels with size $3\times 20\times5\times5$ are shown in Fig. \ref{exp4_a} and Fig. \ref{exp4_b}. The horizontal axis is the output dimension while the vertical axis is the input dimension, with each kernel of size $5\times 5$. To better compare the learned models in different agents under same $r_i$, the results are shown in the order of Agent $2,4,8,6$. Moreover, we select two filters and zoom in for a better comparison between agent $4$ and agent $8$, to show their difference in each kernel.

Firstly, it can be observed in Fig. \ref{exp4_a} that MCE-PL algorithm learns heterogeneous models among agents. The model retention ratios for nodes $2,4,6$ are $0.4, 0.2, 0.1$ respectively. Correspondingly, it can be observed in Fig. \ref{exp4_a} that the sparsity of the learned model becomes larger and the number of non-zero entries becomes smaller. This validates the effectiveness of MCE-PL in node heterogeneity. Secondly, $r_i=0.2$ for both node $4$ and $8$. Compare the learned models of MCE-PL between the two nodes especially shown in the right of Fig. \ref{exp4_a}, certain differences exist between the two partial models. For the $5\time5$ convolutional kernels which are not total zero, the non-zero entries in the two models are different as compared within the blue box. For the output dimension, the fifth to last filter of node $8$ is not exactly zero, while that of node $4$ is all zero as compared within the red box. These differences mainly come from the statistical heterogeneity between the two nodes, validating the effectiveness of MCE-PL for data heterogeneity.
To sum up, the proposed MCE-PL can effectively conduct personalized learning in both data and node heterogeneous conditions.

Then we compare the learned models of the two algorithms in Fig. \ref{exp4_a} and Fig. \ref{exp4_b}. It can be observed the independent updating algorithm \textbf{ind-mask} has more sparsity in the output dimension. Take node $4$ for example. The number of non-zero output dimensions in the selected partial model is $7$ in MCE-PL and $3$ in \textbf{ind-mask}. Such phenomenon mainly comes from the information aggregation procedure in MCE-PL, where the neighboring heterogeneous models have an impact on the local model and the non-zero entries cannot concentrate on only few output dimensions. Such impact may revise the local model and improve the accuracy while reducing the sparsity to some extend. Moreover, as shown in the blue boxes in Fig. \ref{exp4_a} and \ref{exp4_b}, the difference between learned models is much smaller in MCE-PL compared with that in \textbf{ind-mask}, which is resulted from the collaboration among agents in MCE-PL while the \textbf{ind-mask} conducts independent training. 
It should be noted that there may exist multiple binary mask tensors that can achieve good performance due to the non-convexity of the problem. 
In the independent training and without any constraint, the learned subnetworks among agents may share less correlation as shown in Fig. \ref{exp4_b}. On the other hand, the cooperation among agents in MCE-PL equivalently poses the constraint on the subnetworks, which show more similarity.

\subsection{Different Communication Network Topologies}
{In this section, we provide a comprehensive evaluation of the proposed MCE-PL across different communication network topologies. Specifically, we evaluate its testing accuracy under different connectivity probabilities among agents, denoted by $p$, and also consider an extreme case of ring topology. The results, presented in Table \ref{exp9}, show that MCE-PL performs well regardless of the communication network topology, further validating its generalization capability.}

\begin{table}[!htp]
    \centering
    \small
    \caption{{The comparison of averaged testing accuracy under different connectivity probability among agents.}}
    \label{exp9}
    \small
    \setlength{\tabcolsep}{3mm}{
    \begin{tabular}{l|l|l|l|l}
    \hline
    & \multicolumn{1}{l|}{ring} & \multicolumn{1}{l|}{$p=0.3$} & \multicolumn{1}{l|}{$p=0.5$} & \multicolumn{1}{l}{$p=0.7$} \\ \hline
    MCE-PL & {0.6394}     & {0.6489}      & {0.6451}  & {0.6476}\\
    \hline
    \end{tabular}}
\end{table}

\section{Conclusion}\label{conclusion}
This paper proposed a communication-efficient personalized learning algorithm considering both statistical and system heterogeneity, named as MCE-PL. 
Based on the proposed distributed strong lottery ticket hypothesis, the global real-valued parameters for each agent are initialized and fixed,  while each local model is pruned by updating binary mask tensor. To leverage the relationship among agents, each agent's binary mask tensor was transmitted to its neighboring nodes. We designed an aggregation procedure incorporating an aggregation tensor and a personalized fine-tuning step. Finally, we compared the performance of MCE-PL with existing methods designed for heterogeneous conditions. The experimental results validate the superiority of the proposed algorithm. 

{In real-world IoT environments with limited bandwidth and computational constraints, the proposed method is well-adapted with the binary information transmission and heterogeneous pruning ratio, corresponding to computational constraints. Moreover, due to small communication cost and the aggregation procedure only relying on neighboring nodes, MCE-PL has great scalability for large-scale deployments.}
There are still several open research topics. One of them is to consider heterogeneous models with different depth.
Additionally, in practical communication networks, the model retention ratio for each agent cannot be assumed to be known in advance; it needs to be estimated by solving an optimization problem that considers factors like latency or load balancing, while also accounting for the capabilities of each agent.

\numberwithin{equation}{section}
\appendices 
\section{Proof of Theorem \ref{theo2} } \label{appendix:proof Theorem1}
%The proof of Theorem \ref{theo2} is as follows:
By (\ref{eq_theo1}) in Lemma \ref{theo1}, it can be easily known that there exist $\bm{m}_1$ and $\bm{m}_2$ such that the following local properties hold at the same time with probability ${(1-\delta)}^2$:
\begin{align}
&\sup_{\bm{x}\in\mathcal{X}}\big\|f_1(\bm{x})-g_{\bm{m}_1}(\bm{x})\big\|_{\max}\le\varepsilon_1, \notag\\
&\sup_{\bm{x}\in\mathcal{X}}\big\|f_2(\bm{x})-g_{\bm{m}_2}(\bm{x})\big\|_{\max}\le\varepsilon_2. %\inf_{\bm{m}_{2}\in\{0,1\}^{\text{size}(\bm{w})}}
\end{align}
According to the definition of $\|\cdot\|_{\max}$,  we have
\begin{align}\label{eqn:appendix1}
\sup_{\bm{x}\in\mathcal{X}}\big\|f_1(\bm{x})-f_2(\bm{x})-[g_{\bm{m}_1}(\bm{x})-g_{\bm{m}_2}(\bm{x})]\big\|_{\max} \le \varepsilon_1+ \varepsilon_2.
\end{align}
By Assumption \ref{as4}, we have  
\begin{align}
\sup_{\bm{x}\in\mathcal{X}}\|f_1(\bm{x})-f_2(\bm{x})\|_{\max} &\le\alpha_u, \label{eqn:appendix2}\\
\inf_{\bm{x}\in\mathcal{X}}\|f_1(\bm{x})-f_2(\bm{x})\|_{\max} &\ge\alpha_l. \label{eqn:appendix3} \\
\inf_{\bm{x}\in\mathcal{X}}\big\|f_1(\bm{x})-f_2(\bm{x})-[g_{\bm{m}_1}(\bm{x})- &g_{\bm{m}_2}(\bm{x})]\big\|_{\max} \ge 0.   \label{eqn:appendix4}
\end{align}
By combining \eqref{eqn:appendix1} and \eqref{eqn:appendix2}, we have 
\begin{align}\label{eqn:appendix5}
\sup_{\bm{x}\in\mathcal{X}}\big\|g_{\bm{m}_1}(\bm{x})-g_{\bm{m}_2}(\bm{x})\big\|_{\max} \le \varepsilon_1+ \varepsilon_2+\alpha_u.
\end{align}
Then, by combining \eqref{eqn:appendix3}, \eqref{eqn:appendix4}, \eqref{eqn:appendix5}, we have
\begin{align}
&\inf_{\bm{x}\in\mathcal{X}}\big\|g_{\bm{m}_1}(\bm{x})-g_{\bm{m}_2}(\bm{x})\big\|_{\max} \notag \\
& \ge \min\{|\varepsilon_1+ \varepsilon_2-\alpha_l|, |\varepsilon_1+ \varepsilon_2-\alpha_u|, |\alpha_l|\}. 
\end{align}
\noindent This completes the proof.  \hfill $\blacksquare$

\end{document}